\documentclass{article} 
\usepackage[]{iclr2020_conference,times}

\usepackage{amsmath,amsfonts,bm}

\def\eqref#1{equation~\ref{#1}}

\def\1{\bm{1}}

\DeclareMathAlphabet{\mathsfit}{\encodingdefault}{\sfdefault}{m}{sl}
\SetMathAlphabet{\mathsfit}{bold}{\encodingdefault}{\sfdefault}{bx}{n}

\usepackage{hyperref}
\usepackage{url}
\usepackage[pdftex]{graphicx}
\usepackage{subcaption}

\title{State2vec: Off-Policy Successor Features Approximators}

\author{Sephora Madjiheurem \& Laura Toni \\
University College London, UK\\
\texttt{\{sephora.madjiheurem.17,l.toni\}@ucl.ac.uk} \\
}

\newcommand{\ie}{\emph{i.e.,}}
\newcommand{\eg}{\emph{e.g.,}}

\iclrfinalcopy 
\begin{document}

\maketitle
\begin{abstract}
A major challenge in reinforcement learning (RL) is the design of agents that are able to generalize across tasks that share common  dynamics. A viable solution is  meta-reinforcement learning,   which identifies common structures among past tasks to be then generalized to new tasks (meta-test). In meta-training, the RL agent learns state representations that encode  prior information from a set of  tasks, used to generalize the value function approximation. This has been proposed in the literature as   successor representation approximators. While promising, these methods do not generalize  well across optimal policies,  leading to sampling-inefficiency during   meta-test phases. In this paper, we propose \textit{state2vec}, an efficient and low-complexity framework for learning  successor features which \textit{(i)} generalize across policies, \textit{(ii)} ensure sample-efficiency during meta-test. We extend the well known node2vec framework to learn state embeddings that account for the discounted future state transitions in RL. The proposed off-policy state2vec captures the geometry of the underlying state space, making good basis functions for linear value function approximation. 
\end{abstract}

\section{Introduction}
Reinforcement Learning (RL) can be described as the computational approach to learning from interaction. In RL, an agent acts in an environment and receives observations including a numerical reward. The reward is usually a function of the current state (of the environment) and the action taken by the agent. The goal of the agent is to learn how to act, \ie which control strategy (or \textit{policy}) to adopt in specific situations in order to achieve a long term goal (maximizing the long term expected reward). RL problems are typically modelled as Markov Decision Processes (MDPs).

This general formulation allows us to describe a large variety of tasks of practical interest across diverse fields such as game-playing, robotics and traffic control \citep{Mnih2015,Levine2016,ma2018universal,Arel2010traffic}. 
With the rise of deep learning, RL has seen great success in the recent years with artificial agents now outperforming humans in increasingly challenging tasks \citep{Graepel2016,Silver2017}. However, these agents exhibit weak-AI behaviors,  being highly task specific and with limited ability to generalise across   tasks. 
While some tasks may be very different  one from another (\eg, learning how to drive vs learning  to master the game of chess), others do not differ much (\eg, taking the train to work vs taking the train to home). In the latter case, it is clearly desirable for the agent to be able to leverage the knowledge acquired while solving one task to speed up the solving of the other similar task. This ability for autonomous agent to re-use previous knowledge is known as \textit{meta-learning}  \citep{Mehta2008, Schmidhuber1996SimplePO}.  The key challenge is for the agent to  adapt or generalize  to new tasks and new environments that have never been encountered during training time.

In this work, we consider a  meta-reinforcement learning (meta-RL) problem, in which tasks are characterized by the same environment (shared structure) but with the reward function changing arbitrarily across tasks.   
Here, the agent learns at two different time scales: slow meta-learning, exploiting the large past experience accumulated across  tasks (learning of the shared structure), and fast learning on individual tasks.   This enables learning how to quickly adapt to a previously unseen task with little data.
A natural tool  for achieving meta-RL is  the successor representation \citep{Dayan1993}. The latter   decouples the environment from the reward in the value function computation, in such a way that one remain fixed should the other change.  
Several recent works have adopted the successor features approach for meta-RL  across tasks that share common  dynamics \citep{Barreto2017,Barreto2018,borsa2018universal}. These studies are highly promising as they show the ability for autonomous systems to transfer knowledge across tasks.  However, they exhibit  two major limitations: \textit{(i)} the learning of the successor features (meta-training) is expensive 
and \textit{(ii)} the meta-testing is not sample-efficient, especially for tasks that do not share a common or similar optimal policy. The reason for this is that  the learned successor features are heavily dependent on the on-policy experience, requiring a re-training phase for each individual task \citep{lehnert2017advantages}.

In this paper, we address these challenges by developing an off-policy meta-RL algorithm that disentangles task inference and control. The overall goal is to find the optimal balance between data-training and data-testing, by differentiating the data used in the meta-training with respect to the data used to train the policy. We propose \textit{state2vec}, an efficient yet reliable framework for learning the successor features. 
We are interested in learning  features  that capture the underlying geometry of the state space. In particular, we seek the following properties for the features: \textit{(i)} to be learned from data rather than handcrafted $-$ to avoid structural bias, see \citet{Madjiheurem2019};  \textit{(ii)} to be low-dimensional $-$ to ensure a fast adaptation during meta-testing; \textit{(iii)}  to be geometry-aware  rather than task-aware $-$ to generalize across   optimal policies. To learn such features, we extend the well known node2vec algorithm \citep{Grover2016} to  infer graph embeddings  capturing temporal dependencies. In other words, state2vec encodes  states in low-dimensional embeddings, defining the  similarity of states based on the discounted future transitions. Moreover, to ensure off-policy meta-training, we impose that the data used for training is fully exploratory and independent  of any specific task (it is reward agnostic).  
This allows us to use the same representation without any retraining of the features to solve tasks with varying reward functions. In the meta-testing phase, the agent will need to simply learn a task-aware coefficient vector  to derive a value function approximation. The dimensionality of the coefficient vector is imposed by the   embedding dimension, which we constraint to be low to favor   sample-efficiency in the meta-testing.    
We show experimentally that state2vec captures with high accuracy  the structural geometry of the environment while remaining reward agnostic. The experiments also support the intuition that off-policy state2vec representations are robust low dimensional basis functions that allow to approximate well the value function. 

The reminder of this paper is organized as follows: In Section~\ref{sec:background} we formalise the meta-RL scenario in which we are interested and present the background material upon which our work is built. Section~\ref{sec:state2vec} introduces \textit{state2vec}, our novel successor features approximation algorithm. The experimental set-up and results are presented and discussed in Section~\ref{sec:experiments}. We review the relevant literature in Section~\ref{sec:related}, relating and comparing prior works to our approach to meta-RL. Finally, in Section~\ref{sec:conclusion}, we summarise our main contributions and highlight some of the relevant directions for future work.  

\section{Background}
\label{sec:background}

\subsection{Meta-Reinforcement Learning}
In RL, a decision maker, or agent, interacts with an environment by selecting actions with the goal to maximise some long term reward. This is typically modelled as a Markov Decision Process (MDP). 
A discrete MDP is defined as the tuple $M = (S, A, P, R, \gamma)$, where $S$ is a finite set of discrete states, $A$ a finite set of actions, $P$ describes the transition model $-$ with $P(s,a,s')$ giving the probability of visiting   $s'$ from state $s$ once  action $a$ is taken, $R$ describes the reward function and $\gamma \in (0,1]$ defines the discount factor. We consider \textit{finite} MDPs, in which the sets of states, actions, and rewards   have a finite number of elements. The random variable $R$ and $s$ have well defined discrete probability distributions that only depend on the previous state and action, hence having the Markovian property. 
A policy  $\pi $ is a mapping from states to probabilities of selecting each action in $A$. Formally, for a stochastic policy, $\pi(a | s)$ is the probability that the agent takes action $a$ when the agent is in state $s$. Given a policy $\pi$, an action-value function $Q^{\pi}$ is a mapping $S \times A \mapsto \mathbb{R}$ that describes the expected long-term discounted sum of rewards observed when the agent is in a given state $s$, takes action $a$, and follows policy $\pi$ thereafter. Solving an MDP requires to find a policy that defines the optimal action-value function $Q^*$, which satisfies the following constraints:
\begin{equation} \label{eq:bellmansQ}
Q^*(s,a) = R(s,a) + \gamma \sum_{s'\in S} P(s,a,s')  \max_{a'} Q^*(s',a')    
\end{equation}
This recursive equation is known as  \textit{Bellman's equation}. The optimal policy is a  unique solution to the Bellman's equation. In the case of single MDP framework with a  sufficiently small  search space, the optimal solution can be found by dynamic programming, iteratively evaluating the value functions for all states.

The problem becomes more challenging in the case of multiple tasks, which is the focus of this work.  We are interested in the set of MDPs spanned by the tuple $(S, A, P, \gamma)$:  
\begin{equation}\label{eq:multitasksRL}
\mathcal{M} = \{M_1, M_2, \ldots, M_n\}
\end{equation}
with each tasks being a MDP problem $
M_i = (S, A, P, R_i, \gamma)$,  with $ R_i : S \times A \mapsto \mathbb{R}
$. 
In other words, we investigate the meta-learning problem in which  tasks $M_i$  share the same MDP components, except  the reward function, which  is drawn from a common random distribution $M_i\sim p({M_i})$. This is of obvious interest as this formalism  potentially model real life applications. This is the same setting adopted in prior related works \citep{Barreto2017,Barreto2018,borsa2018universal,lehnert2017advantages}. 
In this setting, we are interested in finding out if there is an efficient way of learning the underlying dynamics that is shared across all tasks, such that once this information is known, solving a specific MDP becomes a much easier problem. 

\subsection{Successor Representation}
In order to address the meta learning problem,  we need to decouple   the dynamics of the MDP (common across tasks)  from the reward function (task discriminant) in the value function approximation. This decoupling motivates the adoption of the successor representation, or SR, \citep{Dayan1993}. With the SR, we can factor the action-value function into two independent terms:
\begin{equation}\label{eq:qfunction}
Q^{\pi}(s,a) = \sum_{s'} \Psi^{\pi}(s, a, s') R(s,a),
\end{equation} 
where the SR $\Psi^{\pi}(s, a, s')$ is defined, for $\gamma < 1$, as:
\begin{equation}\label{eq:sr}
 \Psi^{\pi}(s, a, s') = \mathbb{E}_{\pi} \Big[ \sum_{t = 0}^{\infty} \gamma^t \mathbb{I}(s_t = s') | s_0 = s, a_0 = a \Big], 
\end{equation}  
where $\mathbb{I}(s_t = s') = 1$ if  $s_t = s'$  and $0$ otherwise.  

The provided interpretation, is that the SR is a predictive type of representation, which   represents a state action pair as a feature vector $\boldsymbol{\Psi}^{\pi}_{s,a}$ such that, under policy $\pi$, the representation $\boldsymbol{\Psi}^{\pi}_{s,a}$ is similar to the feature vector of successors states.
Computing the action-value function given the SR is computationally easier as it becomes a simple linear computation.  
Furthermore, given the SR, the recomputation of the action value function is robust in changes in the reward function: the new action value function can be quickly recomputed using the current SR.
The SR is therefore a natural tool to consider for transfer in reinforcement learning.

\subsection{Successor Feature}
\citet{Barreto2017} proposed a generalisation of the SR called \textit{successor feature} (SF). They make the assumption that the reward function can be parametrised with
\begin{equation}\label{ed:linearR}
R(s,a) = \phi(s,a)^\top \boldsymbol{w}\,,
\end{equation}

where $\phi(s,a)$ is a feature vector for $(s,a)$ and $ \boldsymbol{w} \in \mathbb{R}^d$ is a vector of weights. Because no assumption is made about $\phi(s,a)$, the reward function could be recovered exactly, hence (\ref{ed:linearR}) is not too restrictive. Under this assumption, the action-value function for the task defined by $\boldsymbol{w}$ can be rewritten as:
\begin{equation}
Q^{\pi}(s,a) = \boldsymbol{\psi^{\pi}}(s, a) ^\top \boldsymbol{w}\,.
\end{equation}
where the successor feature $\boldsymbol\psi^{\pi}(s, a)$ is defined as
\begin{equation}
\boldsymbol\psi^{\pi}(s, a) \doteq \mathbb{E}_{\pi} \Big[ \sum_{t = 0}^{\infty} \gamma^{t-1} \phi_{i+1} |  s_t = s, a_t = a \Big]\,.
\end{equation}

In \citet{Barreto2017}, authors also define the \textit{generalized policy improvement} (GPI) theorem, which shows that given previously computed SF approximation $\hat{\boldsymbol{\psi}}_{\pi_{\boldsymbol{w}}}(s, a)$ for some tasks $M_w \in \mathcal{M}$, the agent can derive a policy $\pi$ for a new task $M_i \in \mathcal{M}$ which is guaranteed to preform at least as well as any previously learned policy.
More specifically, denoting by $\pi_{\boldsymbol{w}}$ the policy followed by the agent in the task ${\boldsymbol{w}}$,  when a new task $\boldsymbol{w}^{\prime}$ is experienced,  the agent will apply the following strategy 
\begin{equation}\label{eq:policy_GPI_2018}
\pi_{\boldsymbol{w}^{\prime}} \in \arg\max_{a}\max_{\boldsymbol{w}} \hat{\boldsymbol{\psi}}(s, a)^\top \boldsymbol{w}\,.
\end{equation}
 In practice, this means that across all the observed tasks, we will consider the best value function when deciding our policy.  The main limitation, however, is the double dependency of the value function on  $\boldsymbol{w}$, which defines the task as well as the policy. This means that if all previous tasks are significantly different than the new task $\boldsymbol{w}^{\prime}$, the derived policy $\pi_{\boldsymbol{w}^{\prime}}$ will be far from the optimal policy for task $\boldsymbol{w}^{\prime}$, and the knowledge of the previous task is not transferable to the new task.   
 
 To disentangle these two contributions, \citet{borsa2018universal} define a \textit{general value function}  and  \textit{universal successor features} respectively by $$Q(s,a;\boldsymbol{w},\pi ) \doteq \boldsymbol{\psi}^{\pi}(s, a)^\top \boldsymbol{w}\,,$$
 $$\boldsymbol{\psi}(s, a; \pi)^\top \doteq \boldsymbol{\psi}^{\pi}(s, a)^\top\,.$$
 
 They further define a \textit{policy embedding}: $e(\pi): (S\times A \rightarrow   \pi) \rightarrow \mathbb{R}^k$. The choice of embedding $e(\pi_{\boldsymbol{z}}) = \boldsymbol{z}$, leads to the following approximation:
\begin{equation}
    \hat{Q}(s,a;\boldsymbol{w},\boldsymbol{z} ) =   \boldsymbol{\psi}(s,a; \boldsymbol{z})^\top \boldsymbol{w}
\end{equation} 
The key novelty is that the  features $ \boldsymbol{\psi}$  are now generalized across policies. 
This leads to the following policy
\begin{equation}\label{eq:policy_USF_2018}
    \pi(s) \in \arg\max_a \max_{ \boldsymbol{z}\in \mathcal{C}}    \boldsymbol{\psi}(s,a; \boldsymbol{z})^\top \boldsymbol{w}^{\prime}.
\end{equation} 
where if $\mathcal{C}$ is the set of tasks used to approximate the SF. It is clear that   $\mathcal{C}= \{\boldsymbol{w}^{\prime} \}$  recovers a universal value function approximator, minimising the value function approximation error. Conversely,  if $\mathcal{C}=\mathcal{M}$, it means that all experienced tasks are considered for the  SF evaluation. This reduces to the GPI of \citep{Barreto2018}, which provides a good generalization across tasks, at a price of a less accurate value function approximation. 

\section{Off-policy Successor Features Approximators}
\label{sec:state2vec}

The main limitation of the proposed methodologies is that they proposed a representation that is transferable only across \textit{similar} policies \citep{lehnert2017advantages}. 
In this work, we are rather interested in proposing a fixed low-dimensional representation, which  \text{(i)} generalizes  across tasks; \text{(ii)} and  is policy independent. Specifically, we target to encapsulate the geometry of the MDP by learning a low-dimensional representation of the states, i.e., \textit{state2vec}. This representation captures the geometry of the problem, which is common across any tasks. At each task, the agent needs to learn only the value function approximation as a function of the low-dimensional state2vec.  
Our work is a special case of (\ref{eq:policy_GPI_2018}) and (\ref{eq:policy_USF_2018}) in which the features are derived from the \textit{state} embedding, aimed at capturing the geometry of the problem, which is the common aspect of the MDPs across different tasks.  Denoting by ${\Psi}(s,a)$ our embeddings we  have 
\begin{equation}\label{eq:policy_s2v}
    \pi(s) \in \arg\max_a\max_{\boldsymbol{\theta_{w^{\prime}}}} \underbrace{\left\{\mathbb{E}_{\pi \in \mathcal{M}}    \left[\boldsymbol{\psi}(s,a; \boldsymbol{\pi})^\top \right]\right\}}_{{\Psi}(s,a)}\boldsymbol{\theta_{w^{\prime}}} .
\end{equation}  

The key difference of our work is that we observe the structure of the problem (captured by the off-policy SF) and learn the best policy under this structure by optimising a weight vector $\boldsymbol{\theta_{w^{\prime}}}$. In other words, we seek to maximise the following value function approximation:
\begin{equation}\label{eq:VF_s2v}
\hat{Q}^{\pi_{\boldsymbol{w^{\prime}}}}(s,a) =  \mathbb{E}_{\pi \in \mathcal{M}}    \left[\boldsymbol{\psi}(s,a; \pi)^\top \right] \boldsymbol{\theta_{w^{\prime}}} = {\Psi}(s,a)^\top \boldsymbol{\theta_{w^{\prime}}}\,,
\end{equation}
where with an abuse of notation, we use $\mathcal{M}$ to define the set of all policies as well as set of all tasks in a fixed environment.

In the following, we define state2vec and describe how we train ${\Psi}$ off-policy to capture the geometry of the MDP.

\subsection{Meta-trainging : State2vec}
It is now clear that the state of the art provides very interesting low-dimensional representation of the MDP problem. However, as discussed, proposed representations still suffer from limited knowledge transferability. Specifically, since the SF are learned while taking decision, it is intrinsically  connected to the task. 

Therefore, we propose to approximate the SF off-policy and to use the same representation across all tasks in $\mathcal{M}$. We proposed an efficient framework for learning continuous feature representations of states, such that, similarly to the SF,  states that are neighbours in time should have similar representation. Our method is directly inspired by \citet{Grover2016}'s \textit{node2vec}, and hence we refer to it as \textit{state2vec}.

State2vec learns state representations based on sample episodes' statistics. It optimises the representations such that states that are successors have similar representation. 
It does so by first collecting a data set $\mathcal{D}_{\pi}$ of $n$ walks $ L = \{(s_0, a_0), (s_1, a_1) \ldots , (s_n, a_n) \}$ by following a sampling strategy $\pi$ for maximum $T$ steps (terminating earlier if it results in an absorbing goal state).
Then, optimise the following objective function:

\begin{equation}
\max_{{\Psi}} \sum_{L \in \mathcal{D}_{\pi}} \sum_{(s,a) \in L} \log Pr(N(s,a) | {\Psi}(s,a)),     
\end{equation}
where $N(s_i,a_i) = \{ (s_{i+1},a_{i+1}),(s_{i+2},a_{i+2}), \ldots,(s_{j},a_{j}), \ldots, (s_{i+T},a_{i+T}) \}$ defines the succession of state action pair $(s_{i},a_{i})$ of size $T$.
Similarly to \citet{Grover2016}, we model the conditional likelihood as 
\begin{equation}
Pr(N(s,a)|{\Psi}(s,a)) = \prod_{(s_j,a_j) \in N(s,a)} Pr(s_j,a_j | {\Psi}(s,a)),
\end{equation}

Unlike the \textit{node2vec} algorithm, we account for the fact that neighbours that are further in time should be further discounted. We do so by modelling the the likelihood of every source-neighbour pair as a  sigmoid weighted by a discount factor:
\begin{equation}
Pr(s_j,a_j|{\Psi}(s,a)) =\gamma^{|-j|} \sigma({\Psi}(s_j,a_j) \cdot {\Psi}(s,a))
\end{equation}

where $\sigma$ denotes the sigmoid function.

\subsection{Meta-testing with state2vec}
Once the state2vec representation are learned, we can use them for solving any task in $\mathcal{M}$ without needing to do any retraining. The solving of task $M_{\boldsymbol{w}} \in \mathcal{M}$ given the structural representation $ \sigma({\Psi}$ reduces to optimising the following value function approximation for the weight vector $\theta_{w}$:
\begin{equation}\label{eq:vf_approx}
\hat{Q}^{\pi_{\boldsymbol{w}}}(s,a) = {\Psi}(s,a)^\top \boldsymbol{\theta_{w}}\,.
\end{equation}

This can be achieve using any parametric RL algorithm, such as fitted Q-learning or LSPI \citep{Riedmiller2005NFQ,Lagoudakis2004}.


\section{Experiments}
\label{sec:experiments}

\subsection{Case study}
\begin{figure}[ht]
\begin{subfigure}{.24\textwidth}
  \centering
  \includegraphics[width=.8\linewidth]{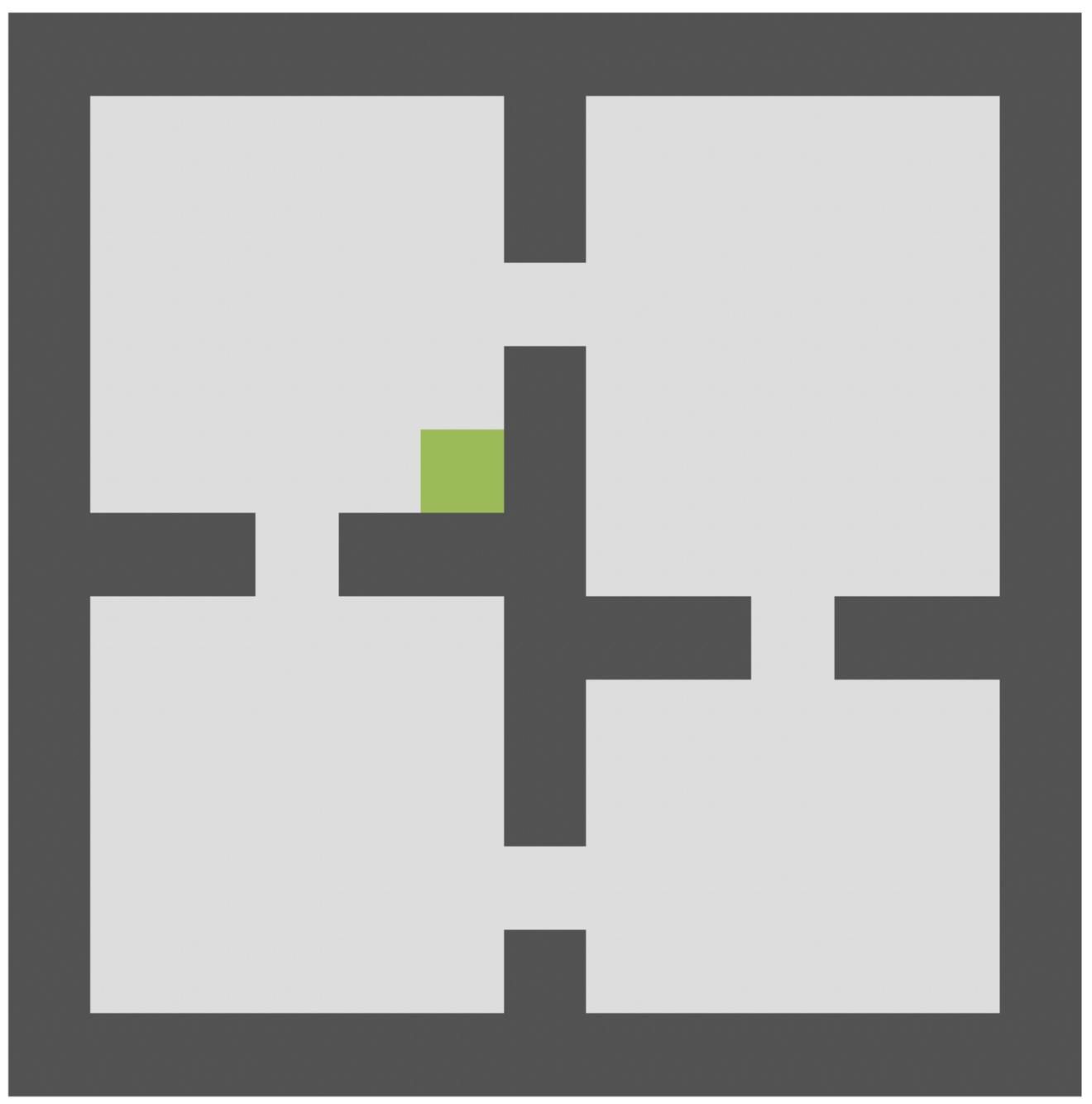}
  \caption{}
  \label{fig:environment1}
\end{subfigure}%
\begin{subfigure}{.24\textwidth}
  \centering
  \includegraphics[width=.8\linewidth]{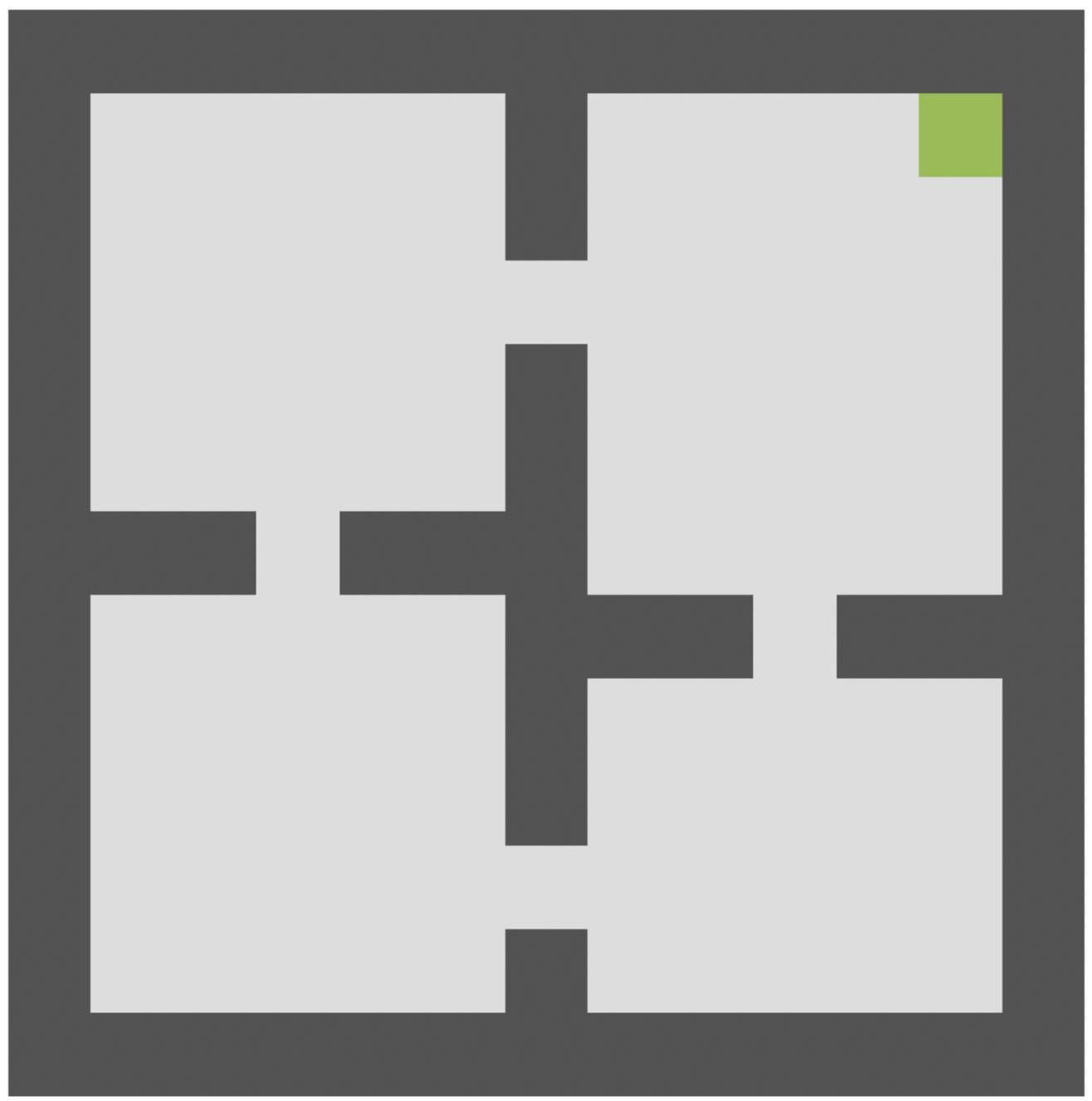}
  \caption{}
  \label{fig:environment2}
\end{subfigure}%
\begin{subfigure}{.24\textwidth}
  \centering
  \includegraphics[width=.8\linewidth]{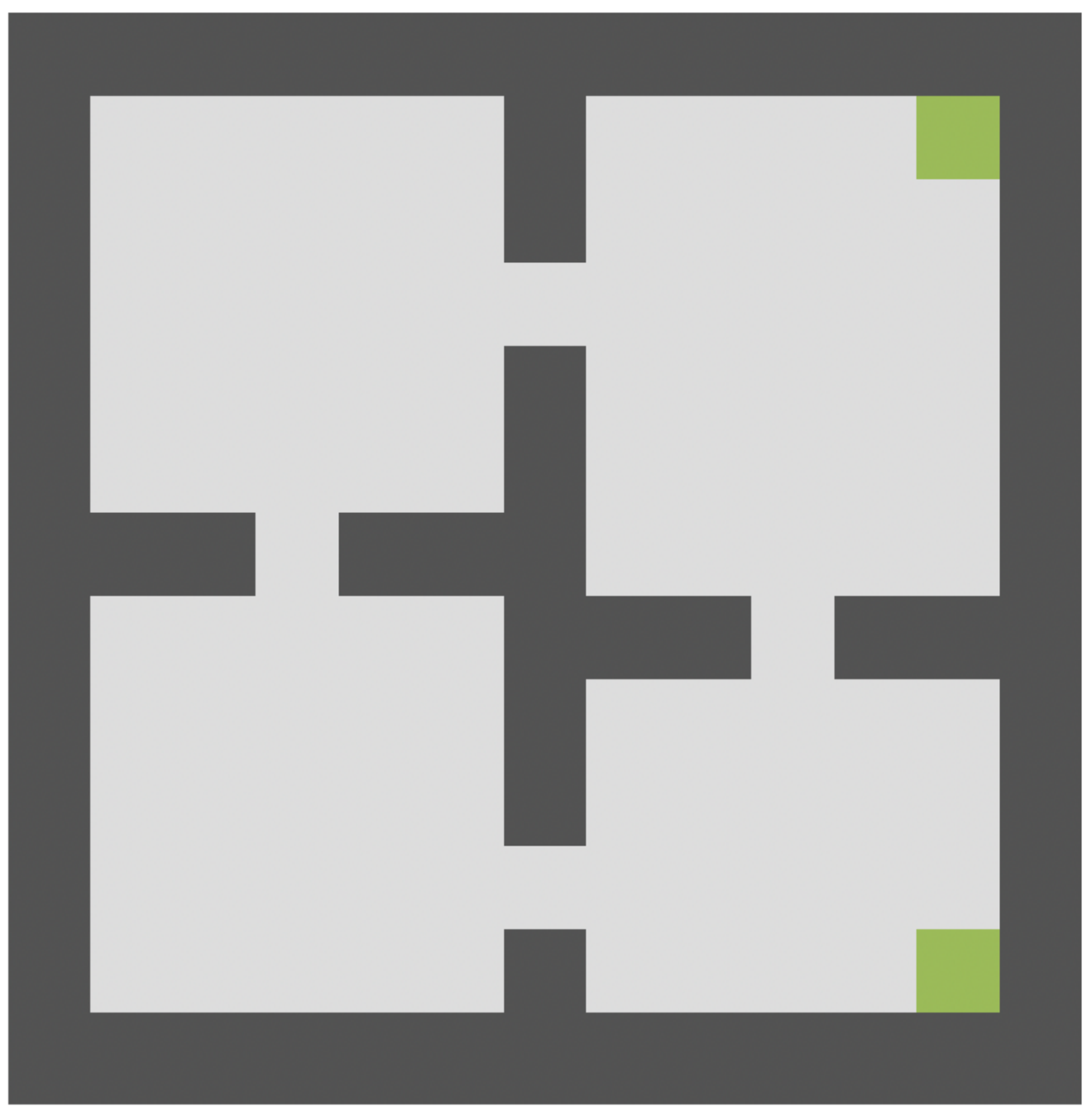}
  \caption{}
  \label{fig:environment3}
\end{subfigure}%
\begin{subfigure}{.24\textwidth}
  \centering
  \includegraphics[width=.8\linewidth]{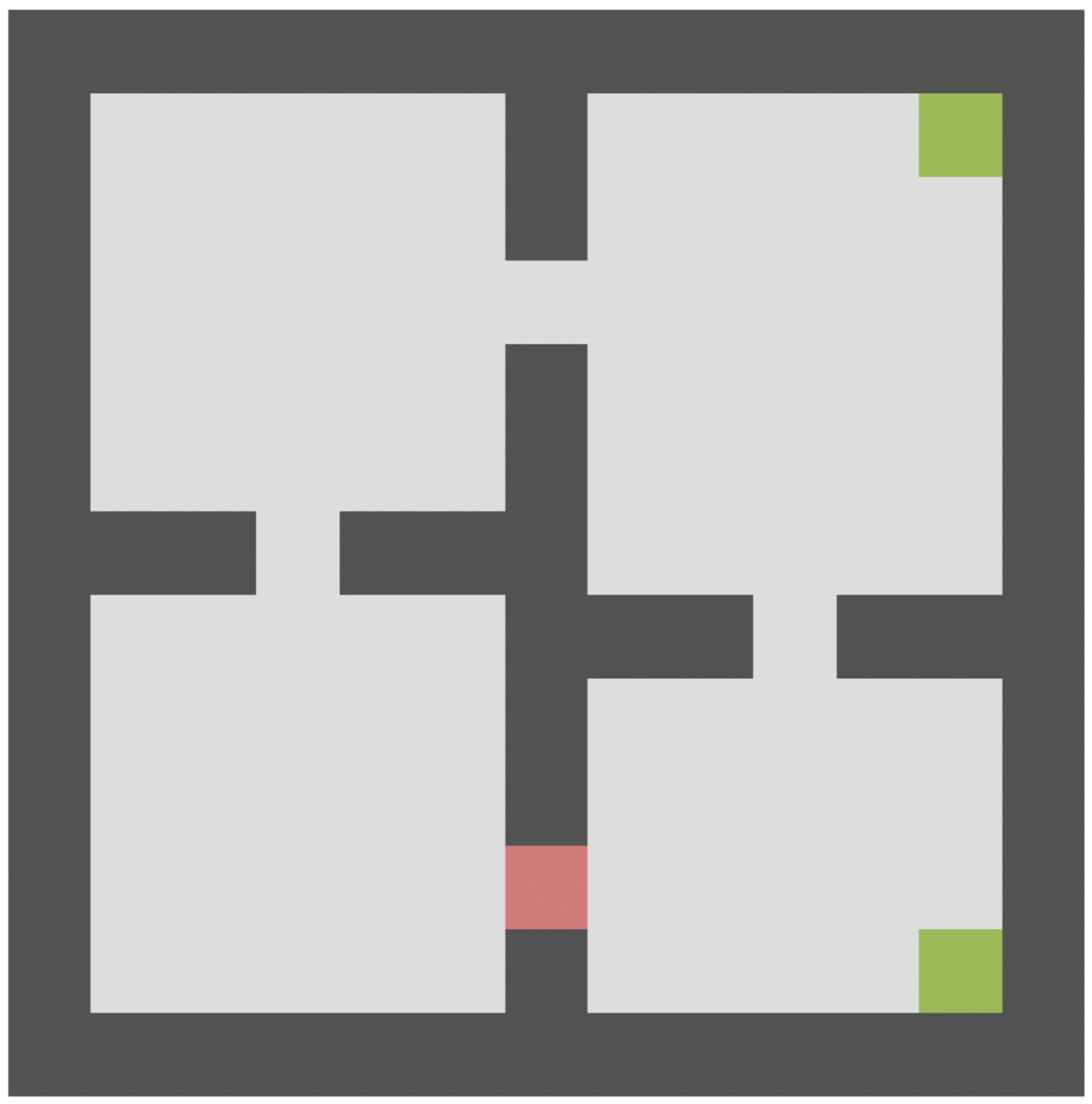}
  \caption{}
  \label{fig:environment4}
\end{subfigure}
\caption{Four-room environment with different configurations.}
\label{fig:fourrooms}
\end{figure}

We consider the four-room domain \citep{Sutton:1999:MSF:319103.319108} shown in Figure~\ref{fig:fourrooms}. It is a two-dimensional space quantized into 169 states, 4 of which are doorways. The agent starts at a random location, and must collect a goal object at a location defined by the task. Depending on the task, the environment also contains ``dangerous'' zones.
The goal object's location is shown in green in Figure~\ref{fig:fourrooms}, while the dangerous states are depicted in red. Collecting an object gives an instantaneous reward of $+100$, and entering a dangerous state gives an instantaneous penalty of $-10$. The the episode terminates when a goal object is  collected or when the agent reaches the maximum of $200$ steps. 

\subsection{Results}
\subsubsection{Meta-training}
\begin{figure}[ht]
\begin{subfigure}{.5\textwidth}
  \centering
  \frame{\includegraphics[width=.8\linewidth]{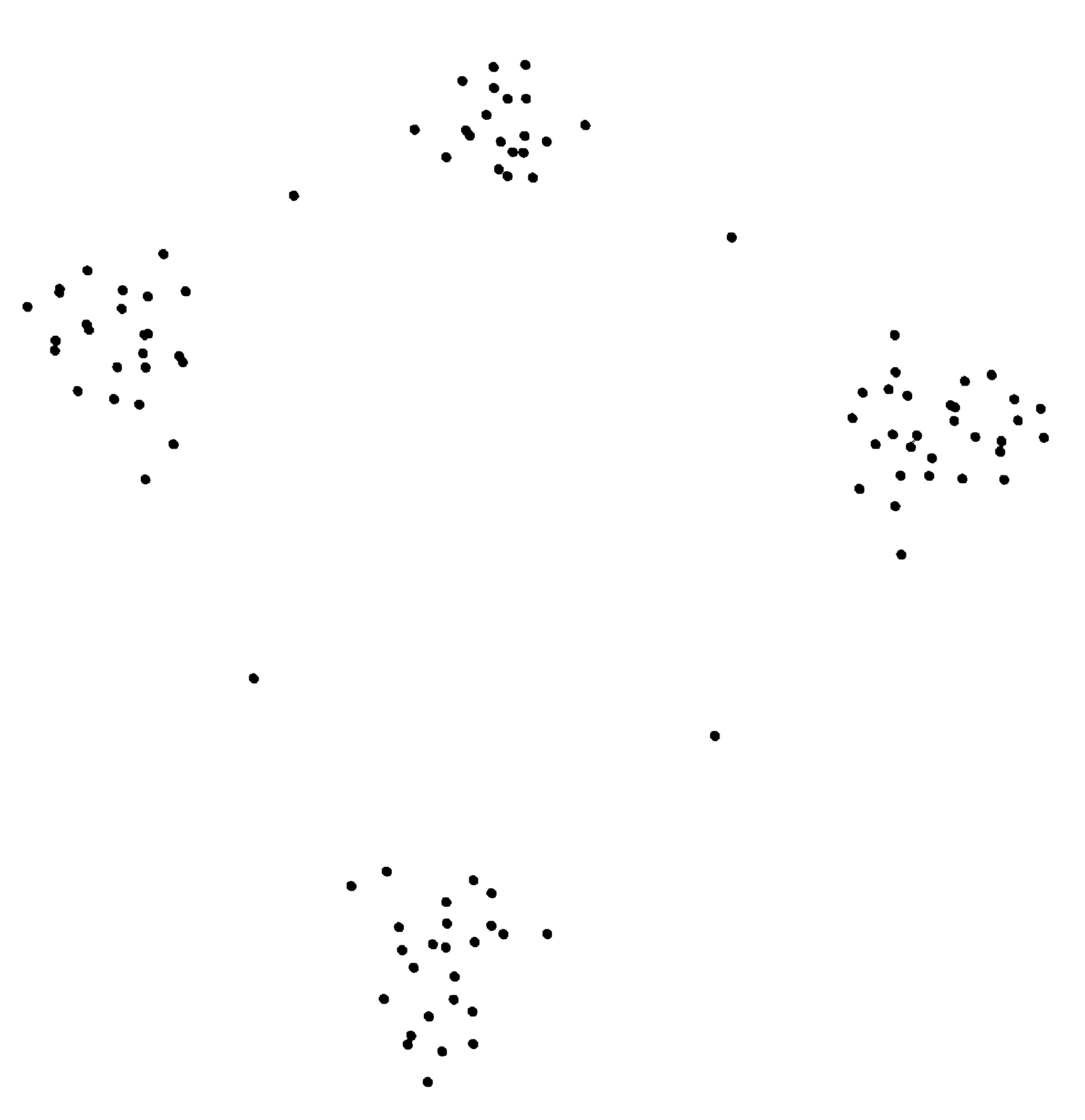}}
\end{subfigure}%
\begin{subfigure}{.5\textwidth}
  \centering
  \frame{\includegraphics[width=.8\linewidth]{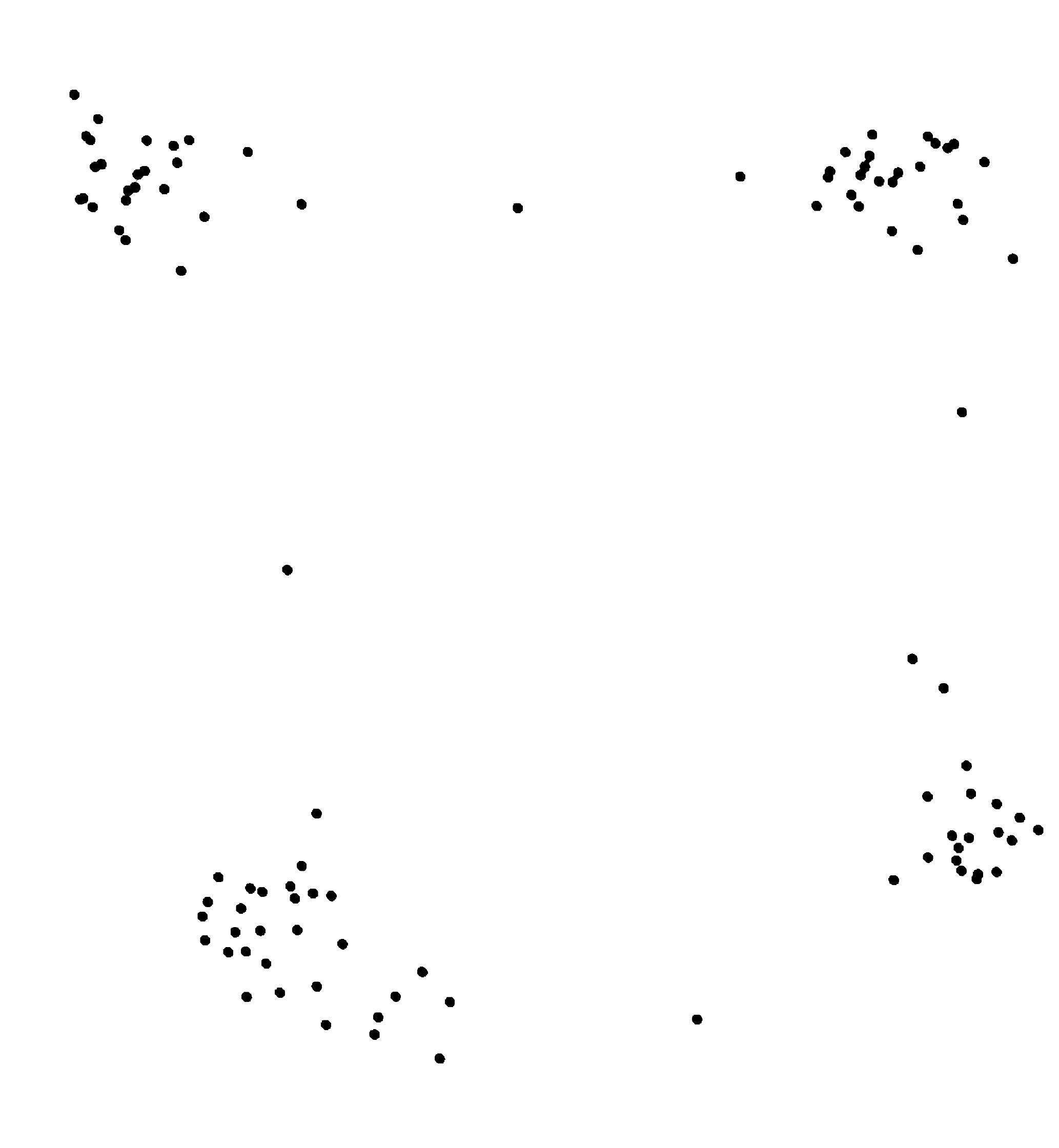}}
\end{subfigure}
\caption{Visualisation of the states representation in feature space (2D PCA projection). \textbf{Left}: the exact successor representation, each vector in the original feature space has dimension 169. \textbf{Right}: the state2vec approximation of dimension 50 in the embedding space. }
\label{fig:pca}
\end{figure}

The meta-training phase is the learning of the state space's geometry by running state2vec. In this feature learning phase, we collect $300$ sample walks of length $100$ and run state2vec with a window size of $50$ and discount factor $\gamma = 0.8$ for varying dimensions $d$.  Figure~\ref{fig:pca} visualises the low dimensional (projection onto the first two principal components) representation of the states in the successor representation and in the state2vec feature spaces. As seen in Figure~\ref{fig:pca}, is a close approximation to the exact successor representation. In both cases, we clearly see that the representations have clustered the states within the same room together, while isolating the doorway states. The learned embedding are shown to preserve the geometry of the state space and identify states that have a special structural role (e.g. doorways).

\subsubsection{Meta-testing}
\begin{figure}
\begin{center}
\includegraphics[width=.8\linewidth]{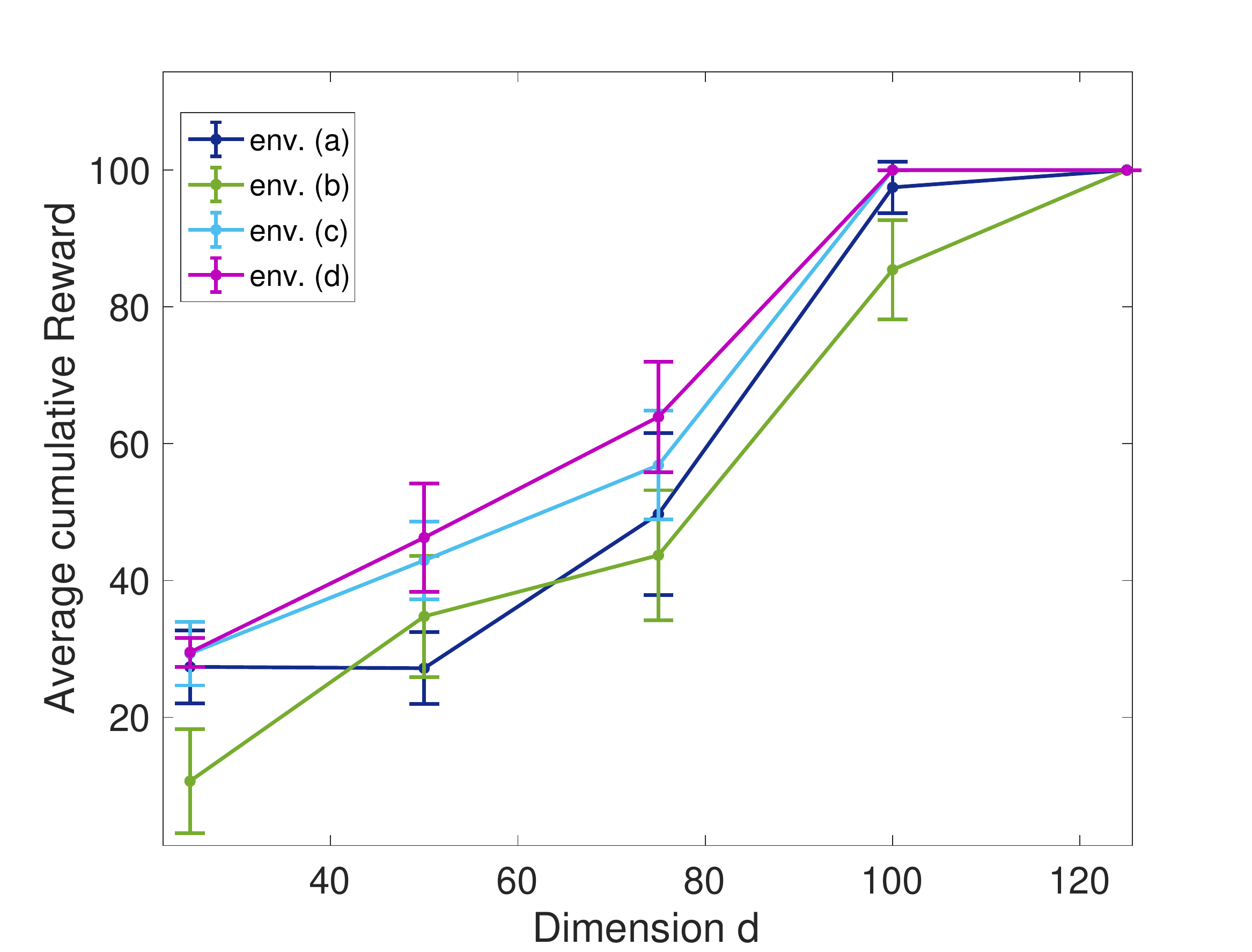}
\caption{Average cumulative reward after meta-testing using pre-trained state2vec for each of the environment in Figure~\ref{fig:fourrooms}.}
\label{fig:dataSize}
\end{center}{}
\end{figure}

\begin{figure}[ht]
\begin{subfigure}{.5\textwidth}
\centering
\includegraphics[width=1.\linewidth]{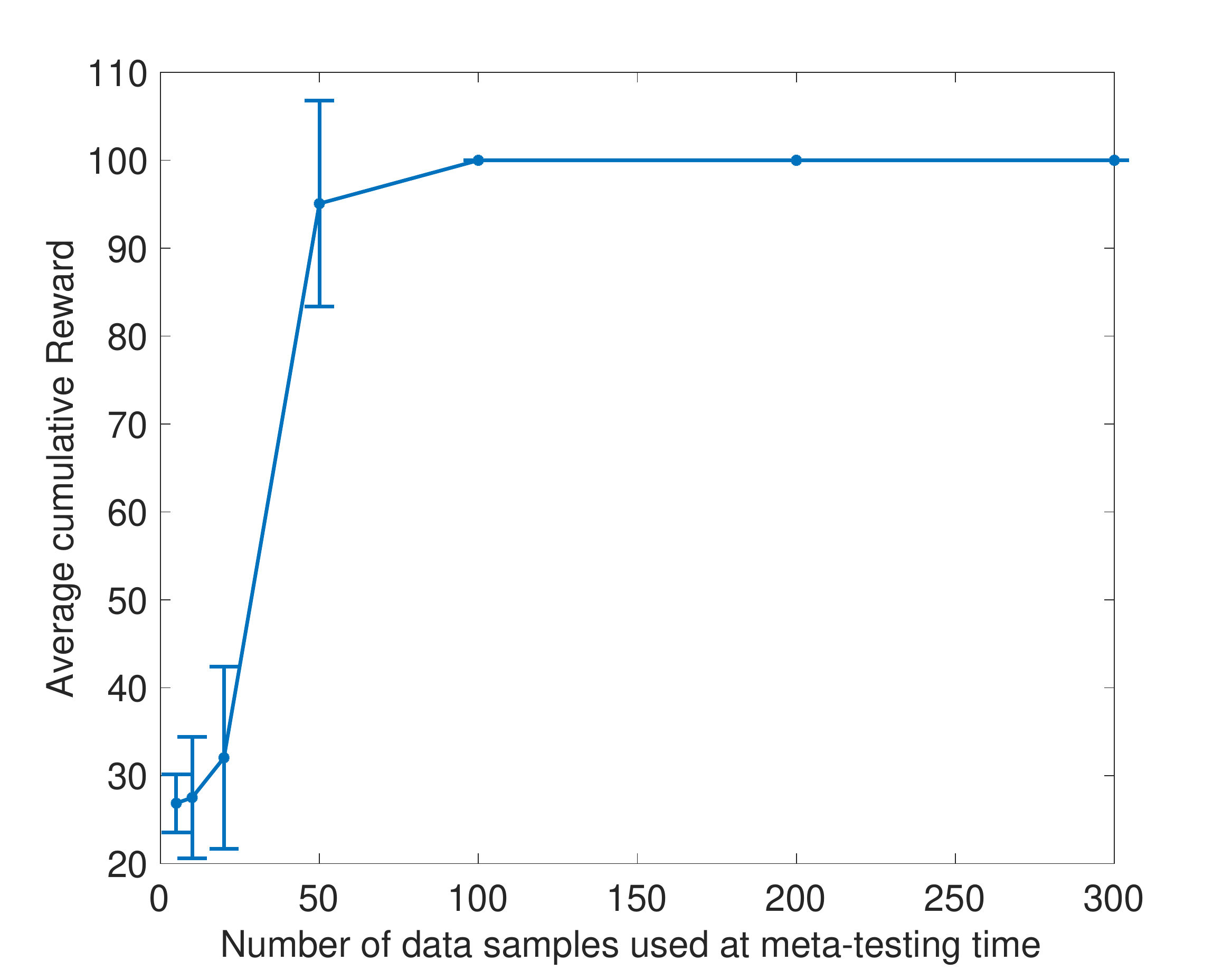}
\caption{Average cumulative reward after meta-testing using pre-trained state2vec (with $d = 100$) for  environment (\ref{fig:environment1}).}
\label{fig:averageReward}
\end{subfigure}
\begin{subfigure}{.5\textwidth}
\centering
\includegraphics[width=1.\linewidth]{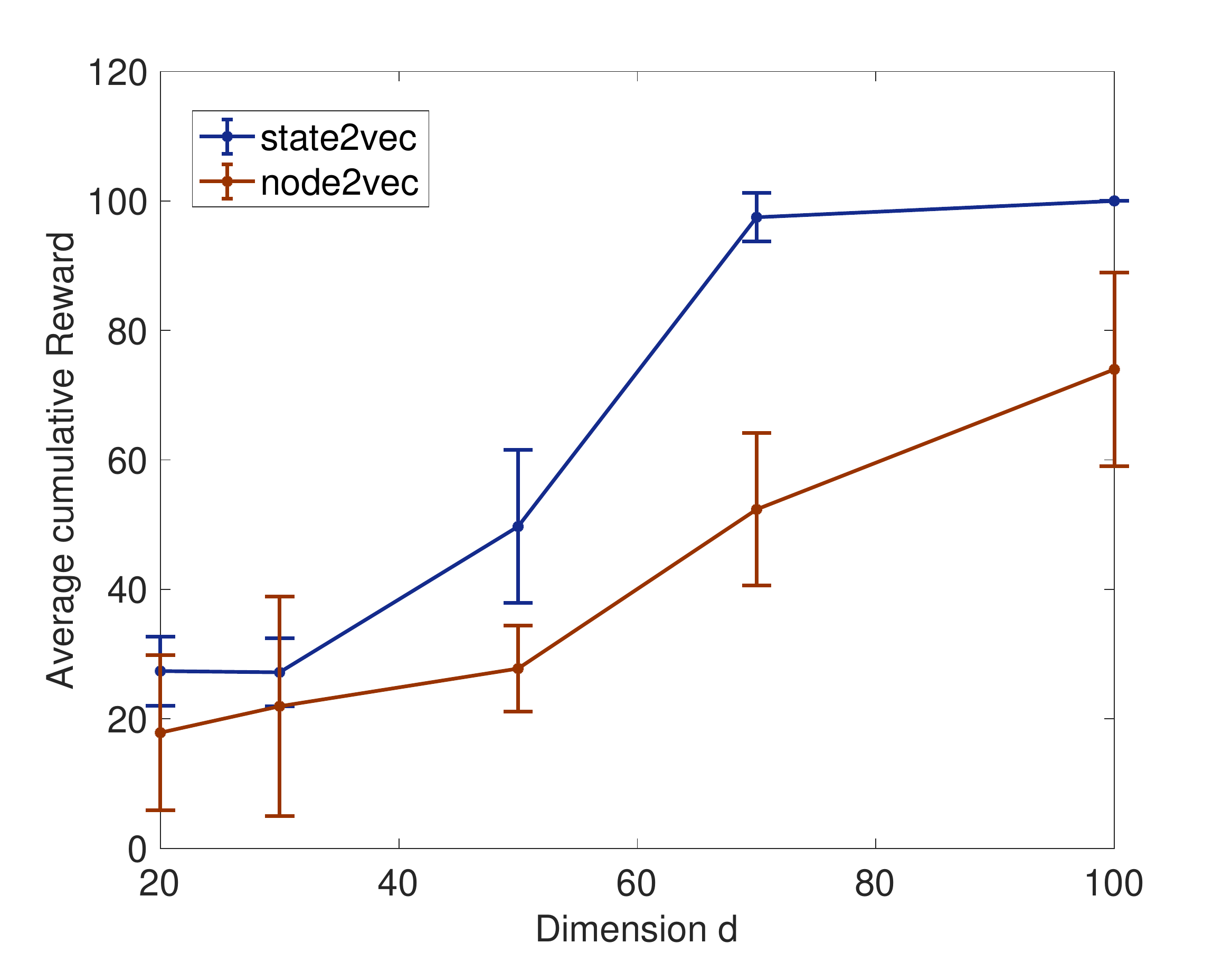}
\caption{Comparison between node2vec and state2vec on environment  (\ref{fig:environment1}) (one goal at the corner).}
\label{fig:n2vVSs2v}
\end{subfigure}
\caption{}
\label{fig:s2vPerformance}
\end{figure}

In meta-testing phase, we use the learned state2vec features to learn the optimal policy of each individual.
We collect sampled realisations of the form $(s, a, s')$ by simulating $50$ episodes of maximum length $200$ (terminating earlier if the goal is reach) and run LSPI \citet{Lagoudakis2004} with state2vec representations as basis vectors to learn the weights $\boldsymbol{\theta_w}$ in \ref{eq:vf_approx}. Figure~\ref{fig:averageReward} shows the performance in terms of average cumulative reward for varying value of $d$. As it can be seen, we are able to achieve strong performance (maximum reward) for all tasks when using the pre-computed state2vec representations of dimensionality $100$ with minimal additional exploration per task. Figure~\ref{fig:dataSize} the performance when we make the size of the data (number of simulated episodes) used at meta-testing vary. We observe a fast reinforcement learning, with optimal policy  learned with only $50$ exploratory episodes collected. These results suggest that information learned during meta-training greatly benefit meta-testing, reducing the need for extensive exploration.

We compare the quality of state2vec embeddings with state of the art low dimensional basis function for linear value function approximation 
Figure~\ref{fig:n2vVSs2v} shows an improved performance of state2vec over node2vec in terms of average cumulative reward. We suspect that the gain in performance comes for the fact that state2vec is design for RL, whereas node2vec is a generic graph embedding algorithm. Specifically, in the objective function, the notion of neighborhood in state2vec is such that further states in time are discounted more than the immediate successors.

\section{Related Work}
\label{sec:related}
The successor feature (SF) as a generalisation of the successor representation \citep{Dayan1993} has been introduced in   \citet{Barreto2017}.  Combining the SF with their \textit{generalized policy improvement} algorithm, they showed how information can be transferred, to some extend, across tasks that share the same dynamics but have different reward functions. 
The initial assumption that rewards can be computed as a linear combination of a set of feature is later relaxed in a follow-up work \citep{Barreto2018}. 
Since  SFs  satisfy a Bellman equation, authors adopt TD-learning to  learn SFs online. The proposed model  generalizes across tasks exploiting the structure of the RL environment. However, the main limitation is that it does not exploit any structural similarity of each single task.  Conversely,   \citet{ma2018universal} introduced the \textit{universal successor representations} (USR) and proposed to model it using a USR approximator. \citet{borsa2018universal}'s \textit{universal successor features approximators} (USFAs) exhibits two types of generalisations: one that exploits the structure in the underlying space of value functions and another that exploits the structure of the RL problem. 

The above works learn the SFs that better generalize the value function approximator across tasks and/or policy. While sharing the same overarching goal, our work 
is conceptually different. We target to learn SFs that encode  the structure of the environment and we ensure off-policy meta-training. In other words,  we aim at learning and leveraging the common structure that is underneath different tasks. The training is performed under off-policy learning. This allows us to decouple the embedding from each single task (either depending on the reward or the optimal policy).  

Some previous works proposed other ways of learning sparse representations that capture the geometry of the state space. \citet{Mahadevan2007} introduced \textit{representation policy iteration}, a framework that jointly learns representations and optimal polices. It builds upon spectral graph theory, and relies on a smoothness assumption of the value function over the state graph. As shown by \citet{Madjiheurem2019}, this assumption does not hold when the estimation of the graph is imperfect, leading to poor value function approximation. In their work, \citet{Madjiheurem2019} adopt \textit{node2vec} as a way of learning the state representations. We have shown that our method state2vec achieves better performance than node2vec. As discussed, the superiority of state2vec for RL probably emerges from the fact that state2vec representations were designed specifically for RL, discounting future occurrences.

\section{Conclusion}
\label{sec:conclusion}
In this work, we focused our effort on the challenging problem of designing RL agents that are able to generalize across tasks that share common dynamics. We considered the meta-reinforcement learning approach, in which the agent, during meta-training, learns a state representation that encodes prior information from the MDP domain, and then leverages this information to solve unseen tasks during meta-testing.  With this goal in mind, we proposed state2vec, an efficient and low-complexity framework for learning state representation. We experimentally showed that state2vec is a good approximation of the successor feature. Additionally, we showed that training the state2vec  off-policy results in embeddings that capture the geometry of the state space and ensure sample-efficiency during meta-testing. The latter simply needs to estimate a low-dimensional reward-aware vector parameter to learn the optimal value function. While promising, our propose method has the limitation that state representations can only be learned for states encountered at meta-training time. Therefore, if not enough exploration can be allocated to meta-training, it is possible that we explore states during meta-testing which have not been seen during meta-training. Consequently, future work should focus on extending state2vec to a model that can generalise across states (meaning we can compute good approximation of the state2vec vector for an unseen state) for the cases where extensive exploration at meta-training time is not feasible. 

We believe the proposed ideas will benefit systems relying on successor representations and will also pave the way for developing meta-reinforcement learning systems that learn representations that are capable of generalising across tasks and are policy-independent, insuring knowledge transferability.  

\bibliography{iclr2020_conference}
\bibliographystyle{iclr2020_conference}


\end{document}